# Data reuse enables cost-efficient randomized trials of medical AI models

Michael Nercessian*, Wenxin Zhang*, Alexander Schubert, Daphne Yang, Maggie Chung, Ahmed Alaa[†], Adam Yala[†]

**Affiliations**
M.N. – UC Berkeley / UCSF Computational Precision Health
W.Z. – UC Berkeley Biostatistics
A.S. – UC Berkeley / UCSF Computational Precision Health
D.Y. – UC Berkeley College of Computing, Data Science, and Society
M.C. – UCSF Department of Radiology and Biomedical Imaging
A.A. – UC Berkeley / UCSF Computational Precision Health
A.Y. – UC Berkeley / UCSF Computational Precision Health

*Joint First
[†]Joint Senior

Corresponding Author: Michael Nercessian
Email: michael.nercessian@berkeley.edu

**Abstract**

Randomized controlled trials (RCTs) are indispensable for establishing the clinical value of medical artificial-intelligence (AI) tools, yet their high cost and long timelines hinder timely validation as new models emerge rapidly. Here, we propose *BRIDGE*, a data-reuse RCT design for AI-based risk models. AI risk models support a broad range of interventions, including screening, treatment selection, and clinical alerts. *BRIDGE* trials recycle participant-level data from completed trials of AI models when legacy and updated models make concordant predictions, thereby reducing the enrollment requirement for subsequent trials. We provide a practical checklist for investigators to assess whether reusing data from previous trials allows for valid causal inference and preserves type I error. Using real-world datasets across breast cancer, cardiovascular disease, and sepsis, we demonstrate concordance between successive AI models, with up to 64.8% overlap in top 5% high-risk cohorts. We then simulate a series of breast cancer screening studies, where our design reduced required enrollment by 46.6%—saving over US$2.8 million—while maintaining 80% power. By transforming trials into adaptive, modular studies, our proposed design makes Level I evidence generation feasible for every model iteration, thereby accelerating cost-effective translation of AI into routine care.

## Introduction

Artificial intelligence (AI) models have the potential to transform patient care by identifying high-risk individuals using high-dimensional data—such as imaging, electronic health records, or time-series data—to personalize screening, prevention, and treatment decisions across a range of diseases, including cancer and heart disease.[1–3] Traditionally, these risk models have been evaluated primarily for their predictive or diagnostic accuracy using retrospective data. However, improved accuracy alone does not necessarily lead to better patient outcomes.[4] To truly assess the clinical utility of AI risk models, there is increasing recognition that we must evaluate their causal impact on outcomes through prospective studies, ideally using the gold standard of randomized controlled trials (RCTs).[5,6] Yet, traditional RCTs, especially for screening interventions, are often highly resource-intensive. They typically require large sample sizes, extended follow-up periods, and complex logistics to detect meaningful differences. This challenge is evident in breast cancer screening, where a recent RCT of risk-based supplemental MRI screening required thousands of participants to demonstrate a reduction in interval cancer.[7] In a decentralized ecosystem where hospitals and academic centers continuously fine-tune or retrain AI models on local data, running a full-scale RCT for every new version is neither financially nor operationally feasible, potentially discouraging future use of the RCT gold standard for evidence generation.

Here, we propose a novel design for prospective trials, *BRIDGE,* that enables scalable and cost-efficient evidence generation for AI models. This design leverages unique characteristics of AI-based risk models used to guide diagnostic or therapeutic decisions that sets them apart from traditional clinical interventions. Unlike conventional interventions, AI risk models do not directly cause changes in patient outcomes—instead, their causal effects are fully mediated through human decisions informed by the models' predictions. Additionally, AI models differ from traditional diagnostic tests in that their predictions can be retrospectively applied to patient data from previous trials. This allows researchers to generate predictions from new models for historical trial populations and assess their potential impact without the need for new patient enrollment. In this sense, data from prior trials already provide partial evidence for newly developed models and can be reused to support their evaluation.

The proposed *BRIDGE* design estimates the causal effect of an AI-driven intervention policy by combining retrospective data from prior trials with prospective enrollment. When the new and legacy models make concordant predictions, data from past trials can be reused. New recruitment is focused on discordant cases, reducing the need for redundant enrollment. Reusing data from previous trials to evaluate new AI models is feasible only under specific technical conditions, which ensure that patients enrolled in the earlier trial would have experienced comparable clinical decisions and outcomes had they been enrolled in the current trial. To assess the validity of data reuse in AI trials, we developed a practical checklist that outlines testable conditions related to the AI model, study population, and outcome measures. These conditions can be evaluated using retrospective data during trial design. This framework is broadly applicable and can be used to evaluate AI-based decision support models across diverse clinical domains.

To assess the efficiency and cost advantages of our adaptive trial design, we carried out case studies in three clinical domains: oncology, cardiology, and sepsis, each representing a common data modality in AI risk prediction—medical imaging (mammograms), electrocardiograms (ECG), and structured electronic health record data (EHR), respectively. These case studies capture common scenarios in AI development: (i) assessing the next generation of an existing model, (ii) externally validating a model developed at a different institution, and (iii) comparing models that rely on different input modalities over time. In all scenarios, the candidate models showed substantial concordance in risk stratification. We quantify how this concordance permits smaller cohorts and markedly lower trial costs. These results demonstrate that prospective trials with data reuse can keep pace with the rapid iteration cycle of clinical AI.

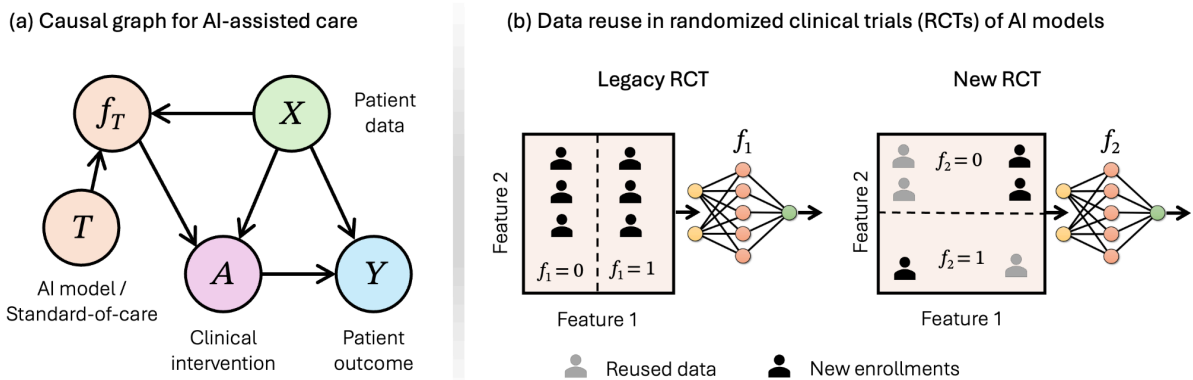

**Figure 1.** Schematic illustrating data reuse in the evaluation of medical AI models. (a) Causal diagram showing the integration of AI models into the clinical workflow, illustrating the relationships between patient features, AI model predictions, clinical interventions, and outcomes. A prospective trial of an AI model aims to evaluate its effect on clinical outcomes under no confounding biases. (b) The *BRIDGE* trial framework enables the reuse of data from previous trials to evaluate new AI models. Reusable data points are determined by the concordance of AI model outputs and the target estimand of the prospective trial.

## Results

*Data Reuse for Cost-Efficient Prospective Trials of Medical AI Models*

In a prospective trial of a medical AI model, the primary objective is to evaluate the effect of deploying the model on future patient outcomes. We focus on AI-based risk screening tools that generate individual risk scores from patient features to guide further diagnostic or therapeutic intervention, denoted by $A \in \{0, 1\}$ (**Figure 1a**). In this setting, a prospective trial of the AI model may involve either randomizing participants to receive AI-assisted care or standard care, or, more pragmatically, using the AI model to identify and enroll only patients flagged as high risk, and randomize the intervention $A = 1$ and standard-of-care within this population. Under the former design, the target estimand is the causal effect of AI-assisted care, similar to a conventional RCT. Under the latter, the target estimand corresponds to the causal effect of administering the intervention $A = 1$ in the population flagged as high-risk by the model, as opposed to the current standard-of-care. Our proposed data-reuse framework accommodates both settings; however, throughout this paper we focus on the latter design.

We consider prospective trials in which the clinical decision to administer the intervention ($A$) follows the AI model recommendation. We represent these decision policies as functions of the patient data $X$, where $f_0$ denotes the standard-of-care decision rule, $f_1$ represents a legacy AI model that has already been evaluated in a prior trial, and $f_2$ corresponds to the updated AI model under evaluation. Each function maps patient features $X$ to a binary decision in $\{0, 1\}$, indicating whether the intervention should be applied. Let $Y$ be a clinically relevant outcome (e.g., cancer detection), where $Y(a)$ denotes the potential outcome that would be observed if the intervention were set to $a \in \{0, 1\}$. The impact of the new AI model can be quantified through the causal effect of the intervention within the population identified as high-risk by the model:

$$\Delta = E[Y(1) - Y(f_0(X)) \mid f_2(X) = 1],$$

tested against a non-inferiority / superiority margin $\delta$. Here, we assume a positive $\Delta$ indicates a positive effect on outcomes under AI-assisted care. When the trial randomizes the AI model in treatment and control arms, then the target estimand is $E[Y(f_2(X))] - E[Y(f_0(X))]$, i.e., the causal effect of deploying the new model within the overall population.

The required sample size $N$ to estimate $\Delta$ can be determined using **Equation 1** in the Methods, based on the predefined superiority margin $\delta$ and standard trial parameters. In a standard trial design, evaluating the first generation of the AI model, $f_1$, requires recruiting $N$ patients based on the formula above to determine whether the model achieves the predefined superiority margin $\delta$. The updated model, $f_2$, trained on additional data or with different methods may become available even before the initial trial for $f_1$ is completed. To generate evidence for $f_2$, the investigator would need to initiate a new trial with a new cohort of $N$ patients and randomizing them to receive care guided by $f_2$. Since AI models are frequently updated or retrained, this approach implies that a new trial must be launched from scratch with each model iteration, which is generally infeasible in terms of time, cost, and scalability.

Our proposed BRIDGE design assumes that the investigator has either previously conducted a randomized trial of an earlier version of the AI model or has access to trial data that supported its validation. To design a new trial evaluating the updated version of the AI model $f_2$, the investigator categorizes each prospective participant into one of two strata based on the agreement between the legacy and updated models:

- **Concordant cases**, identified as high risk by both the legacy and updated models, and
- **Discordant cases**, identified as high risk only by the updated model.

As a new AI model can identify a different high-risk population than a legacy model, the new population (i.e., discordant population) may also have a different treatment effect than the legacy high-risk population. To capture this notion, we stratify the population on the basis of model concordance. Let $C$ denote the patient stratum for whom $f_1(X) = 1$ and $f_2(X) = 1$ in the legacy trial, and $D$ denote the stratum of patients with discordant model outputs $f_1(X) = 0$ and $f_2(X) = 1$. While patients in the discordant stratum are recruited as in a normal RCT, patients in the concordant stratum are recruited only until the required sample size for that stratum is reached after accounting for data reused from concordant patients in the legacy trial **(Figure 1b)**. Consequently, the total sample size $N'$ required to evaluate the new model with data reuse is smaller than the sample size $N$ under a standard design. The corresponding sample size formula is provided in the Methods section.

The degree of concordance determines the extent of reduction in required enrollment that can be achieved through data reuse. In the extreme case where the updated and prior AI models produce fully concordant outputs across the population, no new participants would need to be enrolled, as the trial could rely entirely on previously collected data. Conversely, if the models are entirely discordant, the full original sample size $N$ would be required. Data from the legacy trial can also be reused in future trials if the investigator is interested in other estimands, such as the causal effect of the updated model $E[Y(f_2(X))] - E[Y(f_1(X))]$.

*Validity of Data Reuse in Prospective Trials of Medical AI Models*

The validity of data reuse in our proposed trial design relies on the fundamental assumption of *exchangeability* between historical and prospective trial populations. Specifically, for patients within the same concordance stratum (concordant or discordant model predictions), their potential outcomes under each treatment arm should be statistically equivalent across trials, conditional on observed covariates. In other words, for any given intervention $a$, covariate $x$, AI models $f$ and $f'$, and trials $T$ and $T'$, the following condition holds:

$$E_T[Y(f(X))|X = x] = E_{T'}[Y(f'(X)))|X = x], \text{ if } f(X) = f'(X).$$

| Category | Validity criteria | Data required for testability |
|---|---|---|
| *Population consistency* | <ul><li>Distributions of patient covariates and outcomes are similar across trials within concordant and discordant strata.</li><li>Inclusion/exclusion criteria and recruitment processes are consistent.</li><li>Underlying disease prevalence, risk factor distributions, and treatment responses are stable.</li></ul> | <ul><li>Baseline covariates and outcomes from both recruitment pools; statistical tests for distributional differences.</li><li>Trial protocols, eligibility criteria documentation, recruitment logs, and enrollment patterns.</li><li>Population-level epidemiological data, institutional case mix data, temporal trends in disease incidence.</li></ul> |
| *Outcome consistency* | <ul><li>Outcome definitions and measurement methods are identical.</li></ul> | <ul><li>Outcome adjudication protocols, endpoint definitions, follow-up procedures.</li></ul> |
| *Intervention consistency* | <ul><li>Clinical intervention triggered by legacy and updated AI model recommendations are similar.</li></ul> | <ul><li>Detailed intervention protocols, treatment standardization documentation, care pathway adherence data.</li></ul> |
| *Implementation consistency* | <ul><li>Clinicians comply with AI model recommendations across trials.</li><li>All shared input variables between the legacy and updated AI models are measured identically.</li></ul> | <ul><li>Treatment adherence rates and clinician override patterns, time-to-treatment metrics from both trials.</li><li>Data dictionaries, measurement protocols, laboratory standardization records, imaging acquisition parameters.</li></ul> |

**Table 1.** Checklist of criteria for valid data reuse across randomized controlled trials of medical AI models.

This exchangeability assumption enables valid causal inference while reducing sample size requirements through historical data integration. For this to hold, several key conditions must be met to maintain the validity of the resulting causal inference, which we summarize in **Table 1**. First, the target population must remain consistent between the legacy and current trials to ensure that concordant cases from the earlier study are representative of those who would have been enrolled in the new trial. That is, the population characteristics, inclusion and exclusion criteria, and underlying disease prevalence should be comparable across trials. Second, the outcomes measured in both trials must be consistent. Additionally, the AI model implementations should be similar across trials to avoid introducing confounding effects. Specifically, the models should share the same intended use, operate on similar types of input data measured with comparable technologies, and be embedded within clinical workflows that adhere similarly to the AI recommendations in both trials. Additionally, we assume that clinicians in the trial comply with the AI model recommendations, i.e., no spillover effects and patients with the same prediction always receive the same intervention regardless of the underlying model used.

The conditions described above can be further relaxed when additional patient-level information, such as age, demographics, or other covariates, is available. In such cases, the exchangeability assumptions need only hold within strata defined by these variables, which enables broader applicability for reuse of data. While our primary exposition focuses on RCTs, the principles of data reuse can also be extended to other types of prospective studies, including non-randomized interventional trials.

*Real-World Concordance Between AI Models Across Clinical Domains*

We examined the extent of patient overlap across the predictions of real-world AI models in three clinical domains with diverse data modalities: breast cancer screening (mammograms), cardiovascular risk prediction (ECG waveforms), and sepsis detection (structured EHR data). In each setting, the trial considers an AI model that informs an intervention $a \in \{0, 1\}$ where the standard-of-care is not to administer the intervention (i.e., $a = 0$). In all settings, we analyzed multiple AI models and evaluated their concordance in selecting high-risk patients (**Figure 2**). Summarized results with overlap are shown in **Table 2**. We simulated overlap across a range of thresholds, demonstrating that overlap fraction increases with the size of the high-risk cohort identified (**Figure 3**).

| Clinical domain | Legacy model | New model | Same training dataset? | Same input features? | Same model endpoint? | Top-5% Overlap (%) [95% CI] |
|---|---|---|---|---|---|---|
| *Breast cancer* | AI-Density | Mirai | ✓ | ✓ | — | 2.5 [2.2, 2.9] |
| | ImgOnly DL | Mirai | ✓ | ✓ | ✓ | 46.6 [45.4, 47.8] |
| *Cardiovascular disease* | SEER | S4-ECG | — | ✓ | — | 14.2 [12.2, 16.4] |
| | ResNet | S4-ECG | ✓ | ✓ | ✓ | 49.6 [46.5, 52.7] |
| *Sepsis* | GRU-Dynamic | GRU-Full | ✓ | — | ✓ | 64.8 [64.4, 65.1] |
| | LSTM-Full | GRU-Full | ✓ | ✓ | ✓ | 52.3 [52.0, 52.7] |

**Table 2.** Summarized results for real-world overlap in high-risk cohort identification across clinical domains. Overlap numbers represent the fraction of the top-5% highest-risk cohort identified by the new model that are also identified as high-risk by the legacy model. 95% confidence intervals (CI) are generated using 1000 bootstrap samples.

In our breast cancer case study, we computed three mammogram-based cancer risk scores—Mirai,[2] ImgOnly DL,[8] and AI-Density[9]—on 130,712 screening mammograms from UCSF. These three models were developed by a single team with a single dataset over several years. Mirai and ImgOnly DL were trained on the same modeling endpoint—cancer risk—while AI-Density was trained to predict radiologist BI-RADS density assessments. Among Mirai high-risk exams (top-5%), 46.6% were concordant with ImgOnly DL, and 48.2% were concordant with either ImgOnly DL or AI-Density. The overlap between Mirai and ImgOnly DL was significantly higher than Mirai and AI-Density, reflecting the importance of a common modeling endpoint (46.6% vs 2.5%, p<0.001).

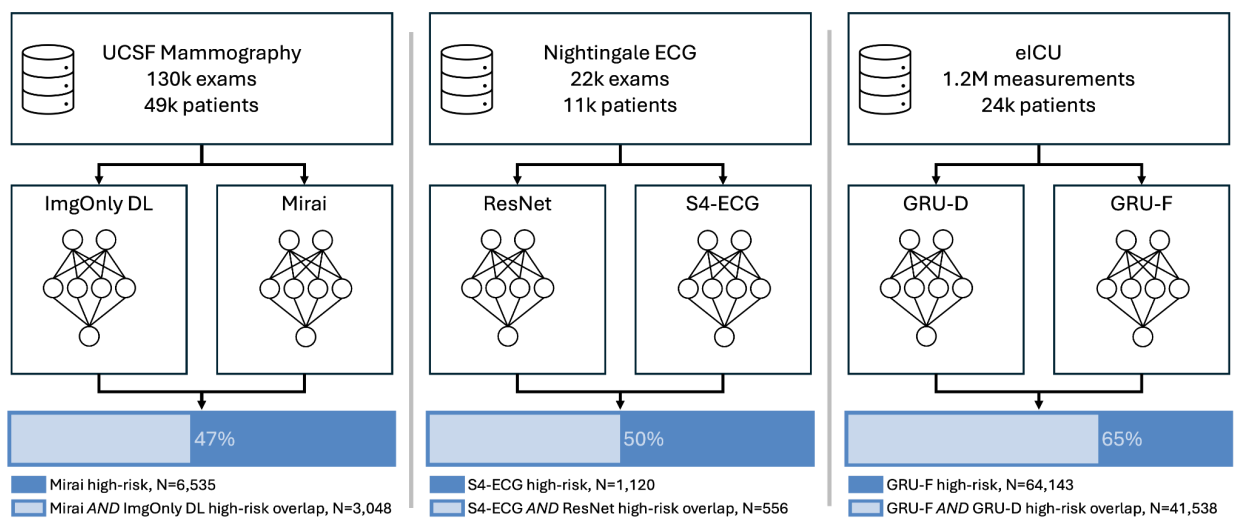

**Figure 2.** Case study workflow and real-world overlap across cancer, cardiovascular disease, and sepsis.

In our cardiovascular disease case study, we computed three ECG-based risk scores—ResNet,[10] S4-ECG,[11,12] and SEER[13]—on 103,952 emergency department ECGs from the Nightingale dataset.[14] We evaluated overlap on a set of 22,400 ECGs from 11,090 patients that were held-out from model development. These models were developed by different groups over time, with different datasets, and reflect the rapid innovation in AI-based cardiovascular risk prediction. ResNet is a deep convolutional model trained to predict 90-day major adverse cardiac events (MACE). S4-ECG is a structured state-space model designed to capture long-range temporal dependencies in ECG waveforms for MACE prediction. SEER is an open-source model trained to estimate long-term cardiovascular mortality risk. Among S4-ECG high-risk patients (top-5%), 49.6% were concordant with ResNet, and 55.7% were concordant with either ResNet or SEER. We again observed significantly higher concordance between models trained on the same endpoint (49.6% vs 14.2%, $p<0.001$).

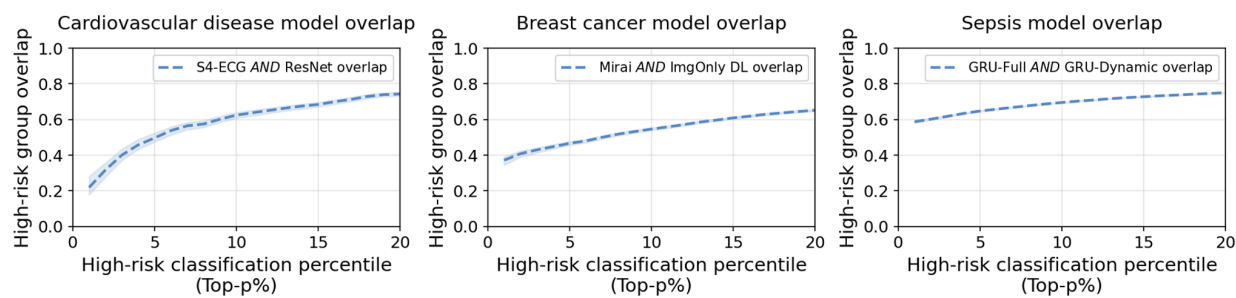

**Figure 3.** Overlap between new and legacy model high-risk cohorts as a function of the percentage of patients classified as high-risk, across cardiovascular disease, breast cancer, and sepsis.

In our sepsis case study, we computed EHR-based risk scores using recurrent neural networks (RNNs) trained and evaluated on 6,417,043 hourly measurements from 123,413 ICU admissions from the eICU dataset. We evaluated overlap on a set of 1,282,848 measurements from 24,683 patients that were held-out from model development. We trained two variations of RNN models, gated recurrent unit (GRU) and long short-term memory (LSTM). We trained versions using varying feature sets: *Dynamic* models incorporated time-varying information such as labs and vitals, while *Full* models combined these temporal features with static clinical features (i.e., demographics). This setup simulates a common development pattern in EHR modeling, where successive models expand input features while maintaining the same prediction target. All models were trained to predict Sepsis-3 onset within six hours, using a 24-hour observational window. Among high-risk patients (top-5%) identified by GRU-Full, 64.8% were concordant with GRU-Dynamic, and 75.2% were concordant with either GRU-Dynamic or LSTM-Dynamic. There was a significant difference in concordance between two different models trained on the same feature set (GRU-Full and LSTM-Full) compared to the same model class trained on different feature sets (GRU-Full and GRU-Dynamic) (52.3% vs 64.8%, p<0.001).

Across all three domains and data modalities, we observed substantial concordance in the identification of high-risk patients between successive AI models, offering an opportunity for accelerated clinical trials with data reuse. These findings suggest that clinical trial data reuse is a broadly applicable opportunity across diverse data types and clinical contexts. Full experimental details, including cohort selection, model implementation, and overlap estimation procedures, are provided in the Methods section.

*Simulated Breast Cancer Trial with BRIDGE*

We simulated a prospective randomized trial evaluating Mirai for allocating supplemental screening MRI in high-risk women. Patients identified as high-risk by Mirai were randomized to receive either supplemental MRI or standard mammography, with the primary endpoint being the reduction in advanced cancer within three years. This design mirrors the landmark DENSE trial,[7] which used breast density to allocate MRI and demonstrated a significant reduction in interval cancers. It is also aligned with more recent trials such as ScreenTrustMRI[15] and the ongoing Mirai-MRI trial,[16] both of which evaluate AI-guided strategies for supplemental screening. We compute sample sizes under (i) a conventional RCT setup, and (ii) with *BRIDGE* data reuse following a completed trial of ImgOnly DL. Our simulated design is shown in **Figure 4**.

We assumed a high-risk breast cancer incidence of 2% and a 30% relative risk reduction with MRI. Under this design, a conventional Mirai trial with no data reuse would require 20,392 patients (4,079 MRI, 16,313 control) to achieve 80% power (type I error rate 0.025, one-tailed). We then simulated a completed prior trial using ImgOnly DL, and identified overlapping high-risk populations. Among Mirai high-risk patients (top-5%), 46.6% (95% CI 45.4%–47.8%) were also flagged by ImgOnly DL. With 100% prior trial completion, *BRIDGE* would enable reuse of 9,503 patient records (1,901 MRI, 7,602 control), cutting prospective enrollment by nearly half. At $1,500 per research MRI, this would save $2,851,500. With 50% prior trial completion, 4,751 records could be reused, saving $1,425,000. These results illustrate how

*BRIDGE* can reduce sample size and trial cost while preserving rigorous evaluation through randomization. The benefit scales with the degree of model concordance and the extent of prior trial completion. While this simulation focuses on breast cancer, the same opportunities apply to trials in cardiovascular disease, sepsis management, and other clinical use-cases.

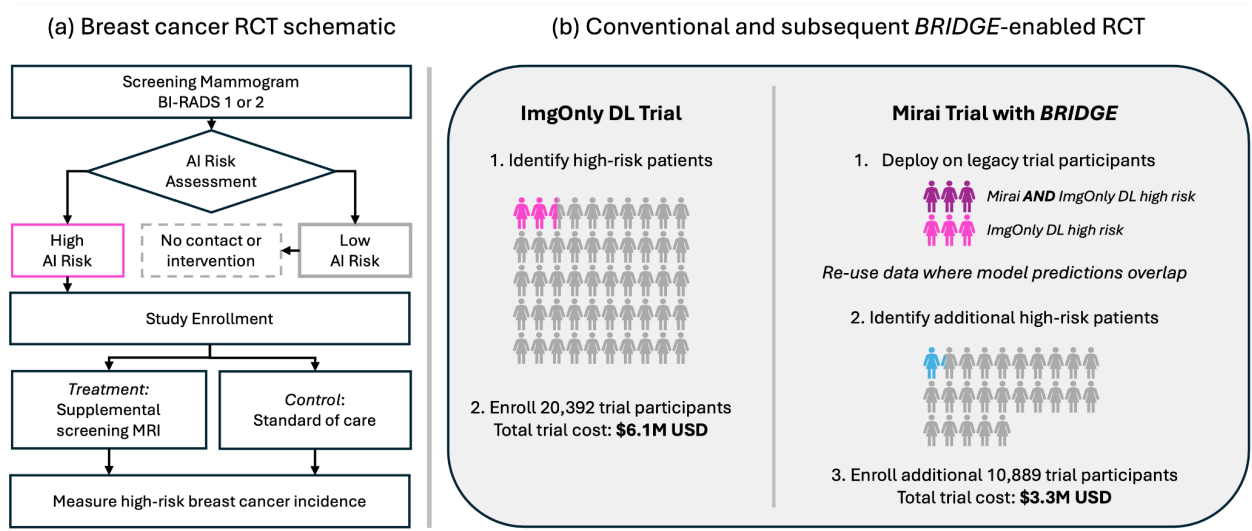

**Figure 4.** (a) Schematic for a simplified breast cancer RCT evaluating AI allocation of supplemental screening MRI to participants who are high-risk. (b) Simulated conventional trial of ImgOnly DL followed by a *BRIDGE*-enabled trial of Mirai, demonstrating data reuse and cost savings.

**Discussion**

The rapid development and deployment of AI models in healthcare has created a fundamental tension between traditional evidence standards and modern software development practices. While clinical interventions typically rely on RCTs for validation, AI systems follow agile development cycles that demand faster iteration and updates. The *BRIDGE* design we propose in this paper addresses this challenge by preserving the core principle of randomization while leveraging existing data to accelerate evidence generation. This approach represents part of an emerging class of study designs that balance scientific rigor with the practical demands of rapidly evolving AI technologies. Recent work has demonstrated that the causal effects of new AI models can be bounded using results from trials of previous models, based on similarities between their predictions, even without prospective testing in new populations.[17] This concept of leveraging historical data for more efficient trial design extends beyond AI applications. Existing Bayesian designs already permit partial reuse of historical data through commensurate or meta-analytic-predictive priors.[18] Similarly, off-policy evaluation in the reinforcement learning literature commonly involves evaluating new decision rules using logged data from previous rules, which is directly relevant to evidence generation for medical AI models, since these models can be viewed as treatment policies.[19]

There is increasing evidence that AI models can improve screening, triage and early-warning tasks across diverse settings from population mammography,[20] to intensive-care sepsis detection.[21] These domains are settings in which we would expect rapid model iteration, where we anticipate our design to deliver the largest gains. In three real-world case studies we observed 46-65% concordance between successive model versions. In our simulated MRI trial, this translated into an absolute enrollment reduction of 9,503 participants and more than US$2.8 million cost savings while maintaining 80% power. A key advantage of our design is that it is model-agnostic, requires no modifications to existing electronic trial-management systems, and interfaces naturally with version-control pipelines routinely used in AI software development. This is in line with the FDA's guidance for predetermined change-control plans (PCCPs) that allow iterative improvement of AI devices without full resubmission.[22] Our design provides a complementary statistical mechanism: while PCCP specifies what will change in AI model and how it will be verified, our design specifies how much data must be prospectively collected to support that verification. Together, they form a pragmatic blueprint for continuous yet rigorous post-market surveillance of evolving AI models.

There are factors that may limit the applicability and use of the *BRIDGE* data-reuse framework. Data drift could violate the exchangeability assumption needed for valid reuse. The relationship between risk factors and outcomes could evolve independently of imaging. For example, if more women who are high-risk based on assessment by Mirai or traditional risk models (e.g., Gail risk model) are now offered prophylactic tamoxifen for breast cancer risk reduction, this would likely lower cancer incidence in the new high-risk cohort compared with an identically defined cohort from a legacy trial. The resulting drop in event rate would violate exchangeability and limit how confidently legacy data can be reused. Upgrades in scanner hardware, changes in acquisition protocols, or demographic shifts in the screened population could result in a legacy dataset becoming unrepresentative of current practice. This could increase the risk of bias emerging from data reuse. As such, detailed records of important variables (e.g. chemoprevention status, MRI scanners in the mammogram-based supplemental screening MRI application) are important for evaluating exchangeability assumptions. Additionally, data reuse could propagate historical biases. For example, if a legacy trial dataset underrepresents racial or socioeconomic minorities, reusing these data without reweighting could carry forward these biases to the new trial.

**Methods**

The local institutional review board approved this Health Insurance Portability and Accountability Act-compliant study and waived the requirement for written informed consent.

*Statistical Methods*

We first define notations and causal estimands. Let $X \in \mathcal{X}$ denote baseline covariates, which can be high dimensional. We use $f: \mathcal{X} \to \{0, 1\}$ to denote an AI model of interest that takes baseline covariates as input and provides a risk classification as output, where: 1 = high risk, 0 = low risk. Let $A \in \{0, 1\}$ denote a binary treatment assignment (1 = treatment, e.g., MRI; 0 = control, e.g., No MRI) informed by the risk classifications, $Y \in \{0, 1\}$ be a binary clinical outcome or event of interest, and $Y(a)$ be the counterfactual outcome under the intervention $a \in \{0, 1\}$.

Suppose we have implemented a trial to evaluate an AI model $f_1$ against the standard-of-care $f_0$. For a new version of the AI model $f_2$, we run a new prospective trial with the target estimand $\Delta = E[Y(1) - Y_{f_0} \mid f_2(X) = 1]$. This represents the average effect of administering the treatment compared to standard-of-care $f_0$ among patients classified with "high risk" by the updated AI model $f_2$. For $X \in \mathcal{X}$, we define $R_i = f_i(X)$, $i \in \{0, 1, 2\}$, for notational brevity. Hence, patient strata with concordant and discordant AI model outputs can be defined through the events $C = \{R_1 = 1, R_2 = 1\}$ and $D = \{R_1 = 0, R_2 = 1\}$, respectively. Without loss of generality, we assume that the standard-of-care does not administer the treatment $A = 1$ for any patient group, hence $R_0 = 0$ for all patients.

To characterize the concordance between AI models, we define $CR_{12} = P(R_1 = 1 \mid R_2 = 1)$ as the fraction of high-risk patients identified by the new AI model who were also flagged as high-risk by the legacy model. Similarly, $CR_{21} = P(R_2 = 1 \mid R_1 = 1)$ is the fraction of high-risk patients under the legacy model still classified as high-risk by the new model. Both $CR_{12}$ and $CR_{21}$ describe the concordance of the AI models in the target population, with $CR_{21} = CR_{12} = 1$ when both models agree in all classifications.

*Identification under data reuse.* Under standard identification assumptions for causal inference, the target estimand $\Delta = E[Y(1) - Y_{f_0} \mid f_2(X) = 1]$ can be written as:

$$\Delta = CR_{12}(E[Y \mid A = 1, C] - E[Y \mid A = 0, C]) + (1 - CR_{12})(E[Y \mid A = 1, D] - E[Y \mid A = 0, D])$$

A complete derivation is provided in the Appendix. The decomposition above shows that the estimand $\Delta$ can be identified using two estimable conditional expectations within the population where the legacy and new AI models are concordant (first term), and the population where the models are discordant (second term). The weight of each term depends on the concordance rate $CR_{12}$. Intuitively, if all patients flagged as high-risk by the new model were also identified by the old model (i.e., $CR_{12} = 1$), then evidence on the new model can be generated by estimating the effect of the treatment $A$ in the stratum $C$ within the legacy trial. If the models are fully discordant

(i.e., $CR_{12} = 0$), then the investigator will need to start a new prospective trial from scratch, recruiting patients in the stratum $D$. In general, when the legacy and new models are partially concordant, the investigator can reuse data points from stratum $C$ and recruit new patients in stratum $D$. The terms above can be estimated within each stratum using simple difference-in-means methods applied to new and reused data points, as long as the data exchangeability assumptions listed earlier are satisfied.

*Sample size calculation under data reuse.* We assume that the investigator has run a legacy trial for model $f_1$ with a sample size of $N_1$ and a treatment allocation ratio $k_1$. To compute the sample size required for a new prospective trial of $f_2$, the investigator needs to specify the target event rates $p_C(a)$ and $p_D(a)$ under intervention $a \in \{0, 1\}$, as well as the treatment allocation ratio $k_2$. Given these parameters, the sample size required to evaluate $f_2$ using a standard design is:[23]

$$N_2 = \frac{(k+1)(z_\alpha + z_\beta)^2}{(\Delta - \delta)^2} \times [CR_{12} \times V_C(k) + (1 - CR_{12}) \times V_D(k)],$$

**Equation 1**

where $V_i(k) = \frac{1}{k} p_i(1)(1 - p_i(1)) + p_i(0)(1 - p_i(0))$, for $i \in \{C, D\}$, depends on the variance of outcomes in the treatment and control arms within the concordant and discordant strata. With data reuse, the required sample size for evaluating $f_2$ is given by

$$N'_2 = N_2(1 - CR_{12}) + max\{0, N_2 CR_{12} \frac{k_2}{k_2+1} - N_1 CR_{21} \frac{k_1}{k_1+1}\}$$

$$+ max\{0, N_2 CR_{12} \frac{1}{k_2+1} - N_1 CR_{21} \frac{1}{k_1+1}\}.$$

The derivation of the above formula is provided in the Appendix. When the legacy and new trials have identical treatment allocation ratios and sample sizes under a standard design (i.e., $N_1 = N_2$ and $k_1 = k_2$), the formula simplifies to $N'_2 = N_2((1 - CR_{12}) + max\{0, CR_{12} - CR_{21}\})$. When the patients identified as high-risk by the new model include none who were classified as high-risk by the legacy model ($CR_{12} = 0$), no data can be reused from the legacy trial, and thus $N'_2 = N_2$. Conversely, when the legacy and new models are identical ($CR_{12} = CR_{21} = 1$), all data are reusable, yielding $N'_2 = 0$, and no new trial is required.

*Implementation of BRIDGE and Details of Case Studies*

To demonstrate the real-world applicability of *BRIDGE*, we performed experiments across three diverse application areas, namely cancer, cardiovascular disease, and sepsis. In each area, we outline datasets and demonstrate overlap across relevant AI models. We also outline a realistic clinical trial in the cancer case study.

*Cancer Case Study*

To study AI-based breast cancer risk models, we collected 130,712 screening mammograms performed at UCSF between 2006 and 2023, encompassing 49,166 unique patients. Cancer outcomes were extracted by parsing free-text pathology notes in the electronic medical record using Strata,[24] a recently validated large language model (LLM) tool. For each mammogram, we computed three AI-driven risk scores: Mirai,[2] ImgOnly DL,[8] and AI-Density.[9] Mirai is a deep learning model that jointly encodes all four standard screening views to predict five-year breast cancer risk. ImgOnly DL, a precursor to Mirai, processes each view independently and was trained on a subset of the data used to develop Mirai. AI-Density, in turn, is an earlier model designed to replicate radiologist-assigned breast density scores—a commonly used image-derived proxy for cancer risk. All models were applied without modification using their respective open-source implementations.

The simulated breast cancer trial design is a simplified RCT design modeled after the Mirai-MRI trial. We assumed a high-risk breast cancer incidence of 2% and a 30% relative risk reduction with MRI. We have published a sample size calculator for *BRIDGE*-enabled trials at https://bridge-trials.org/ which was used for the sample size calculation. For all case studies, hypothesis testing for overlap in model predictions was performed using McNemar's test for paired nominal data. 95% confidence intervals were calculated using 1000 bootstrap samples.

*Cardiovascular Disease Case Study*

Our cardiovascular case study focuses on the prediction of MACE using ECG waveform data. Rapid and accurate assessment of patients at risk of such events is crucial, especially in the Emergency Department (ED), where a high volume of patients present with symptoms suggestive of acute cardiac events. Identifying high-risk patients based on initial clinical presentation alone is challenging; prompt intervention, however, is essential to prevent severe cardiac outcomes.[25] This clinical necessity has driven the development of predictive risk models. Traditional clinical scoring systems such as the Thrombolysis in Myocardial Infarction (TIMI),[26] Global Registry of Acute Coronary Events (GRACE),[27] and History, ECG, Age, Risk factors, and Troponin (HEART)[28] have become important components of current clinical guidelines for timely recognition of patients at risk of Acute Coronary Syndrome (ACS).[29] In recent years, AI-driven prediction models employing residual neural network (ResNet) architectures to assess cardiovascular risk based on ECG waveforms have shown strong performance compared to traditional risk scores (e.g. SEER,[13] ROMIAE study,[30] ECG-MACE[31]). Innovation in AI-based ECG prediction technologies continues to advance rapidly.

We evaluated AI-based MACE prediction models using data publicly provided by Nightingale Open Science,[14] comprising 103,952 ECGs of 44,713 patients who had an ECG taken in the emergency department (ED) of a prominent urban hospital, spanning from January 2010 to May

2015. In the dataset, MACE cases are labeled following the clinical literature, which defines these events using the EHR in a way that shows good agreement with expert judgment after chart review.[32] These events encompass three categories: (i) delayed diagnosis and treatment of blockage with resultant heart muscle damage, verified by troponin levels; (ii) malignant arrhythmias, identified through diagnosis codes and resuscitation procedures; and (iii) mortality, ascertained via the Social Security Death Index data. We trained an 18-layer ResNet model and a structured state-space sequence model, as initially proposed by Gu et al.[11] and subsequently refined in ECG-specific applications[12,33] (S4-ECG). Additionally, we evaluated the open-source Stanford Estimator of Electrocardiogram Risk[13] (SEER), which predicts long-term cardiovascular mortality risk.

*Sepsis Case Study*

Our final case study focused on sepsis prediction. Early identification and treatment of sepsis have motivated extensive development of predictive risk models.[34] Traditional clinical scoring systems, such as the Sequential Organ Failure Assessment (SOFA) and Systemic Inflammatory Response Syndrome (SIRS) criteria, underpin clinical guidelines[35] aimed at timely sepsis detection. Advanced machine learning approaches, including gradient boosting trees and RNNs, have demonstrated improved predictive accuracy over traditional scoring methods,[34,36,37] and several companies including Epic have commercialized AI-driven sepsis prediction systems. Sepsis tools are frequently retrained to address data drift,[38] and the development of improved sepsis risk models remains a prominent area of research.

We developed and validated AI-based sepsis prediction models using data from the eICU Collaborative Research Database (eICU),[39] comprising ICU electronic health records from over 200,000 critical care unit encounters for over 139,000 unique patients admitted between 2014 and 2015. Models predicted the onset of Sepsis-3 within 6 hours given 24 hours of ICU data. Sepsis-3 defines sepsis onset as an increase in SOFA-score of >= 2 points within a window of 48 hours before and 24 hours after a Suspicion of Infection, which occurs when concomitant orders of antibiotics and microbiological samples are taken. Following the dataset construction criteria from prior open-source work,[40] we excluded eICU data with an invalid admission, length of stay, clinical measurements, age, lack of sepsis onset, and a sepsis onset before the 6-hour window. We obtained a sepsis cohort of 6,417,043 hourly measurements from 123,413 patient stays, among which 5,638 (4.6%) satisfied the Sepsis-3 criteria.[35] This dataset was split into 64% training, 16% validation, and 20% test sets. For each hourly timepoint, we extracted 4 static features (age at hospital admission, patient weight, patient height, and sex) and 48 dynamic (time-varying) features (7 vital sign metrics and 41 laboratory measurements). The prediction of Sepsis-3 was performed at the hourly level. We trained two variations of RNN models: GRU and LSTM. Both GRU and LSTM models leveraged temporal dependencies in patient data to predict the onset of sepsis within the defined future window. All models were implemented using the

YAIB framework.[40] These models could be used to allocate high-risk patients to enhanced monitoring and intervention with the goal of preventing severe sepsis or septic shock.

**Data availability**

Raw outputs and binarized predictions for all case studies are available upon request. Data for the sepsis case study is available through the eICU Collaborative Research Database: https://eicu-crd.mit.edu/. Data for the cardiovascular disease case study is available through Nightingale Open Science: https://docs.ngsci.org/. Data for the breast cancer case study is not available for release due to institutional policy to protect patient privacy.

**Code availability**

Code and model weights to run the breast cancer models is available at https://github.com/yala/Mirai, and code to run SEER is available at https://github.com/weston100/AI-ECG. Code and model weights for other models used in this study is available upon request. Code used for the case study analyses will be published upon publication.

**Acknowledgements**

The authors thank Dr. Michael Oberst (Johns Hopkins University) for his valuable comments and feedback on this manuscript.

**Conflicts of interest**

None.

**Contributions**

A.Y. and A.A. conceived the study. M.N. led software development and case studies; W.Z. led statistical methods; M.N., A.S. and D.Y. executed case studies; M.N., A.S. and A.A. prepared figures; M.C. provided expert clinical input. All authors contributed to manuscript drafting. A.Y. and A.A. are co-senior authors; M.N. and W.Z. contributed equally.

# Appendix

*Identification of the causal estimand under data reuse*

Assumption 1. Stationarity: Trial indicator is independent of potential outcomes: $T \perp\!\!\!\perp Y(a) \mid X, R_0, R_1, R_2$.

Assumption 2. Exchangeability: For any treatment $a \in \{0, 1\}$, $A \perp\!\!\!\perp Y(a) \mid R_0, R_1, R_2$.

Assumption 3. Positivity: For any treatment $a \in \{0, 1\}$ and $r_1 \in \{0, 1\}$,
$P(A = a \mid R_0 = 0, R_1 = r_1, R_2 = 1) > 0$.

Assumption 4. $E_T[Y(f(X)) \mid X = x] = E_{T'}[Y(f'(X))) \mid X = x]$, if $f(X) = f'(X)$.

Using the law of total probability, the estimand $\Delta$ can be written as:

$$\Delta = E[Y(1) - Y_{f_0} \mid R_2 = 1]$$

$$= P(R_0 = 0 \mid R_2 = 1)\{E[Y(1) \mid R_0 = 0, R_2 = 1] - E[Y_{f_0} \mid R_0 = 0, R_2 = 1]\}$$

$$+ P(R_0 = 1 \mid R_2 = 1)\{E[Y(1) \mid R_0 = 1, R_2 = 1] - E[Y_{f_0} \mid R_0 = 1, R_2 = 1]\}$$

We assume the treatment is not administered under standard of care, i.e., $P(R_0 = 0 \mid R_2 = 1) = 1$. Accordingly, $\Delta$ can be expressed as follows after applying the law of total probability again:

$$\Delta = P(R_1 = 1 \mid R_2 = 1) E[Y(1) \mid R_1 = 1, R_2 = 1] + P(R_1 = 0 \mid R_2 = 1) E[Y(1) \mid R_1 = 0, R_2 = 1]$$

$$- P(R_1 = 1 \mid R_2 = 1) E[Y_{f_0} \mid R_1 = 1, R_2 = 1] - P(R_1 = 0 \mid R_2 = 1) E[Y_{f_0} \mid R_1 = 0, R_2 = 1],$$

which can be expressed in terms of the concordance rate $CR_{12} = P(R_1 = 1 \mid R_2 = 1)$ as follows:

$$\Delta = CR_{12} E[Y(1) \mid R_1 = 1, R_2 = 1] + (1 - CR_{12}) E[Y(1) \mid R_1 = 0, R_2 = 1]$$

$$- CR_{12} E[Y_{f_0} \mid R_1 = 1, R_2 = 1] - (1 - CR_{12}) E[Y_{f_0} \mid R_1 = 0, R_2 = 1].$$

Using the identifiability assumptions above, we have $Y_{f_0} = Y \mid A = 0$ and $Y(1) = Y \mid A = 1$. Thus,

$$\Delta = CR_{12} E[Y \mid A = 1, R_1 = 1, R_2 = 1] + (1 - CR_{12}) E[Y \mid A = 1, R_1 = 0, R_2 = 1]$$

$$- CR_{12} E[Y \mid A = 0, R_1 = 1, R_2 = 1] - (1 - CR_{12}) E[Y \mid A = 0, R_1 = 0, R_2 = 1],$$

which can be written in in terms of strata $C = \{R_1 = 1, R_2 = 1\}$ and $D = \{R_1 = 0, R_2 = 1\}$ as:

$$\Delta = CR_{12} (E[Y \mid A = 1, C] - E[Y \mid A = 0, C]) + (1 - CR_{12})(E[Y \mid A = 1, D] - E[Y \mid A = 0, D]).$$

Following assumption 1, the conditional expectation $E[Y \mid A = a, C]$, $a \in \{0, 1\}$, is equivalent in legacy and new trials, hence it can be identified from reusable data in $C$.

*Sample size calculation under data reuse*

The point estimator of Δ can be written as

$$\hat{\Delta} = CR_{12}[\hat{p}_C(1) - \hat{p}_C(0)] + (1 - CR_{12})[\hat{p}_D(1) - \hat{p}_D(0)].$$

The variance of the point estimator is given by:

$$v^2 = CR_{12}^2 \left[\frac{p_C(1)(1-p_C(1))}{n_C(1)} + \frac{p_C(0)(1-p_C(0))}{n_C(0)}\right] + (1 - CR_{12})^2 \left[\frac{p_D(1)(1-p_D(1))}{n_D(1)} + \frac{p_D(0)(1-p_D(0))}{n_D(0)}\right]$$

We assume the sample proportion in each strata matches the distribution $P(R_1 = r_1 | R_2 = r_2)$, and $n_i(1) = k\, n_i(0)$ for $i \in \{C, D\}$. For an overall sample size $N_2$, we have that

$$v^2 = CR_{12}^2 \left[\frac{p_C(1)(1-p_C(1))}{N_2 CR_{12} \frac{k}{k+1}} + \frac{p_C(0)(1-p_C(0))}{N_2 CR_{12} \frac{1}{k+1}}\right] + (1 - CR_{12})^2 \left[\frac{p_D(1)(1-p_D(1))}{N_2 (1-CR_{12}) \frac{k}{k+1}} + \frac{p_D(0)(1-p_D(0))}{N_2 (1-CR_{12}) \frac{1}{k+1}}\right]$$

$$= CR_{12} \frac{k+1}{N_2} \left[\frac{1}{k} p_C(1)(1 - p_C(1)) + p_C(0)(1 - p_C(0))\right]$$

$$+ (1 - CR_{12}) \frac{k+1}{N_2} \left[\frac{1}{k} p_D(1)(1 - p_D(1)) + p_D(0)(1 - p_D(0))\right].$$

The required sample size to estimate Δ can be determined based on the predefined superiority margin δ. The null hypothesis typically posits that the causal effect $\Delta \leq \delta$ (one-sided test), indicating no clinically meaningful benefit of the AI model. The sample size needed to control type I error α and achieve power $1 - \beta$ is given by solving for $v^2 = (\Delta - \delta)^2/(z_\alpha + z_\beta)^2$, i.e.,

$$N_2 = \frac{(k+1)(z_\alpha + z_\beta)^2}{(\Delta - \delta)^2} \times [CR_{12} \times V_C(k) + (1 - CR_{12}) \times V_D(k)],$$

where $V_i(k) = \frac{1}{k} p_i(1)(1 - p_i(1)) + p_i(0)(1 - p_i(0))$, for $i \in \{C, D\}$. The total sample size for treatment randomization within each stratum is,

$$n_C = N_2 CR_{12} = n_C(1) + n_C(0), \quad n_C(1) = \frac{k}{k+1} n_C, \quad n_C(0) = \frac{1}{k+1} n_C; \text{ and}$$

$$n_D = N_2 (1 - CR_{12}) = n_D(1) + n_D(0), \quad n_D(1) = \frac{k}{k+1} n_D, \quad n_D(0) = \frac{1}{k+1} n_D;$$

The reusable samples in both trial arms can be obtained by subtracting concordant data points from the legacy trial as follows

$$n_C(1) = max\{0, N_2 CR_{12} \frac{k_2}{k_2+1} - N_1 CR_{21} \frac{k_1}{k_1+1}\}, \text{ and}$$

$$n_C(0) = max\{0, N_2 CR_{12} \frac{1}{k_2+1} - N_1 CR_{21} \frac{1}{k_1+1}\},$$

Whereas the discordant samples are given by $N_2(1 - CR_{12})$. Hence the sample size under data reuse is given by:

$$N'_2 = N_2(1 - CR_{12}) + max\{0, N_2 CR_{12} \frac{k_2}{k_2+1} - N_1 CR_{21} \frac{k_1}{k_1+1}\}$$

$$+ max\{0, N_2 CR_{12} \frac{1}{k_2+1} - N_1 CR_{21} \frac{1}{k_1+1}\}.$$